\documentclass{article}
\usepackage{spconf,amsmath,graphicx,amsfonts, hyperref}

\usepackage{enumitem}
\usepackage{xcolor}
\setlist{nosep, leftmargin=14pt}

\usepackage{mwe} % to get dummy images

\title{Automatic Report Generation for Histopathology images using pre-trained Vision Transformers and BERT}
\name{Saurav Sengupta, Donald E. Brown}
\address{School of Data Science, University of Virginia, Charlottesville, USA}
\begin{document}
%\ninept
%
\maketitle
\begin{abstract}
Deep learning for histopathology has been successfully used for disease classification, image segmentation and more. However, combining image and text modalities using current state-of-the-art (SOTA) methods has been a challenge due to the high resolution of histopathology images. Automatic report generation for histopathology images is one such challenge. In this work, we show that using an existing pre-trained Vision Transformer (ViT) to encode 4096x4096 sized patches of the Whole Slide Image (WSI) and a pre-trained  Bidirectional Encoder Representations from Transformers (BERT) model for language modeling-based decoder for report generation, we can build a performant and portable report generation mechanism that takes into account the whole high resolution image. Our method allows us to not only generate and evaluate captions that describe the image, but also helps us classify the image into tissue types and the gender of the patient as well. Our best performing model achieves a 89.52\% accuracy in Tissue Type classification with a BLEU-4 score of 0.12 in our caption generation task.
% which gender the tissue came from, that is, all the information that is available in the Genotype-Tissue Expression (GTEx) portal\footnote{https://www.gtexportal.org/home/histologyPage}.
\end{abstract}
\section{Introduction}
\label{sec:intro}

High resolution histopathology slides are a rich resource of information that current deep learning methods are able to exploit for various use cases like disease classification, cell segmentation and outcome prediction. However, as the images are very high resolution, usually in the range of 150,000x150,000px, they often require non-trivial modifications to existing SOTA deep learning architectures to be used successfully. The most common method for handling these high resolution images is to patch the bigger image into smaller sized images that can be fed into Convolutional Neural Networks. For example, in a classification setting, this often works as a multiple instance learning problem, where each patch is given the same overall image label. A potential drawback to this is that patching can lead to removal of overall context from the WSI that the model might need to learn to make the correct decision, unless handled properly.

Automatic report generation for histopathology images is an area of research where we can modify existing SOTA image captioning architectures to fit researchers needs. Image captioning for histopathology helps us combine two rich sources of information, that is, high resolution WSIs and associated diagnostic reports that describe features of the image. In clinical settings, automatic report generation has been successfully used for X-ray images and claim to reduce the burden for radiologists by assisting them in describing the image \cite{park2020feature}. Other use cases for automated image captioning in medical images can be image retrieval, as generated reports could be part of a searchable database, and encouraging standardized clinical ontologies by using words from a standard vocabulary to describe similar things. Therefore, automated image captioning for histopathology can be similarly useful for a wide variety of tasks that can assist physicians and radiologists in their tasks.

\begin{figure*}[htb]
\begin{minipage}[b]{1.0\linewidth}
  \centering
  \centerline{\includegraphics[width=0.7\textwidth]{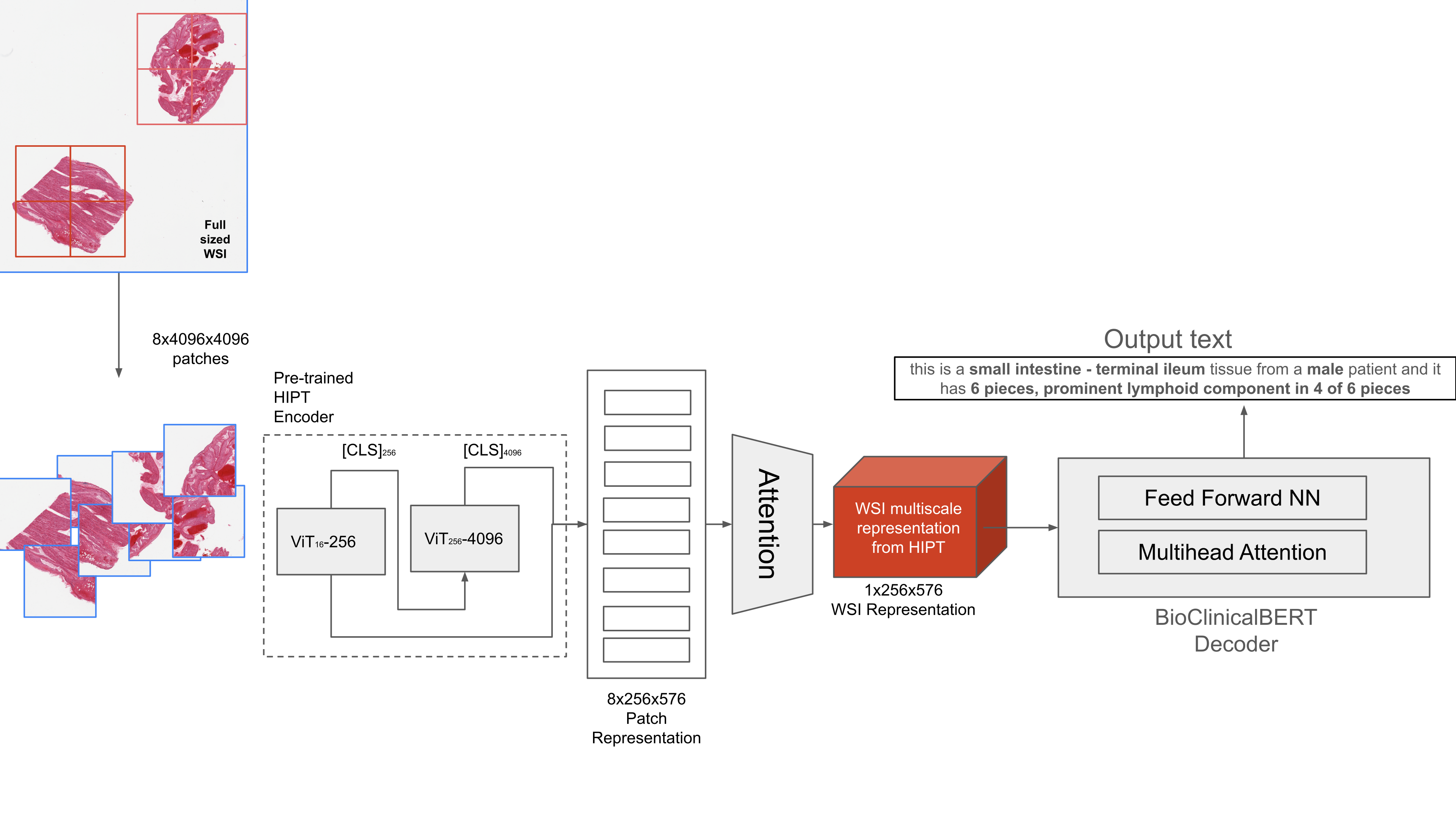}}
\end{minipage}
\caption{Method Overview}
\label{fig:overview}
\end{figure*}

\section{Related Work}
\label{sec:rw}

Current research for histopathological image captioning focuses on Convolutional Neural Network (CNN) based encoder and Recurrent Neural Network (RNN) based decoder architectures \cite{zhang2020evaluating, gamper2021multiple, tsuneki2022inference}. This is inspired by Show, attend and tell paper, that in particular has the capability of using the attention mechanism to focus on certain areas of the image to generate captions \cite{xu2015show}. 

% Attention values for explanation is an existing area of research as explained by \cite{haab2022attention} and is often a requirement to validate the correctness of the output in clinical settings where model reliability and interpretability is of the utmost importance.

Using ImageNet pre-trained CNN-based encoders to encode smaller sized patches of the high resolution WSI has been successfully used in a variety of ways to essentially reduce the size of large dimensional WSIs to smaller and computationally manageable representations \cite{chen2021annotation}. In recent years,  a self-supervised Vision Transformer (ViT) based image representation learning mechanism called Hierarchical Image Pyramid Transformer (HIPT) has been proposed \cite{chen2022scaling}. The self-supervised pre-training leverages DINO (distillation with no labels) at two levels, 256x256 sized patches and 4096x4096 sized patches. The authors show that this can then be leveraged for further downstream tasks like disease sub-typing and survival prediction\cite{caron2021emerging}.

A recent work uses a two-step process in which they first encode all patches of a WSI using a triplet loss based convolutional autoencoder and use the features from the bottleneck layer to cluster the patches into $k \in [1,2,3..7]$ clusters \cite{zhang2020evaluating}. In the second step they randomly sample the patches from each cluster, use a ImageNet pretrained ResNet-18 to extract $N$-dimensional features for the $k$-patches, then use attention pooling to reduce $k \times N$ dimensional feature vector to $1 \times N$ and then feed into a LSTM decoder to generate captions.
More recently, Gamper and Rajpoot \cite{gamper2021multiple} describe the ARCH dataset which contains histopathology images extracted from textbooks and their associated descriptions, which they use for caption generation based pre-training task to generate an encoder that when used for downstream tasks like multiple instance learning shows promising results compared to other pre-trained encoders. 
% But as noted in Tsuneki and Kanavati in \cite{tsuneki2022inference}, these images, as they are curated from textbooks and research articles, can be of mixed quality, magnifications, and resolutions that while useful, does not really solve the problem for existing high resolution WSIs being generated in hospital systems everywhere.
In Tsuneki and Kanavati \cite{tsuneki2022inference}, the authors use high resolution WSIs from a Japanese hospital system and associated translated text reports, for their automated captioning system. They use EfficientNetB3 \cite{tan2019efficientnet} and DenseNet121 \cite{huang2017densely} pretrained on ImageNet dataset and extract features from the penultimate layer for 300x300 patches extracted from the WSI. They then use global average pooling and 3x3 average pooling to reduce the feature sizes and feed them into an RNN based decoder for generating their captions.

% \vspace{-0.2cm}
BERT is a transformer model pre-trained on a large corpus of unlabeled text data that is able to learn bidirectional representations for text and has been successfully used for various downstream tasks in like Question Answering (QA), Named Entity Recognition (NER) and Natural Language Inference (NLI) \cite{devlin2019google}. BioBERT and ClinicalBERT, trained on medical research articles and clinical notes data respectively, performed better on clinical downstream tasks than just BERT \cite{alsentzer2019publicly}. 

% \vspace{-0.2cm}
More recently, pre-trained transformers have been successfully used for optical character recognition (OCR), that is, converting text in images to machine-readable text \cite{li2023trocr}. The authors were able to outperform SOTA approaches for OCR using pre-trained Vision Transformers and transformer-based language models. The authors utilize the image representations from the vision transformers and, along with the context generated before, use it to predict the next tokens. A `[BOS]' and `[EOS]' tokens are appended at the beginning and end of the ground truth tokens. Note that using `[BOS]' token shifts the sequence to the right by one place and is used to indicate start of generation.

Motivated by the success of pre-trained transformer models for such diverse downstream tasks, we propose a method that uses HIPT to encode WSIs that captures multi-level representations, and a BERT based decoder that is able to utilize powerful text representations to generate descriptions of the WSI.

% In this paper, we use the same dataset as \cite{zhang2020evaluating} with their available train/val/test splits\footnote{https://github.com/zhangrenyuuchicago/PathCap}, specifically the same test data for comparison purposes and use pre-trained $\mathrm{ViT_{256}}$-16 representations from \cite{chen2022scaling} and an ViT based encoder and borrowing the LSTM based caption generator from \cite{xu2015show}. We show that we achieve comparable results on the same test set as used by \cite{zhang2020evaluating} and further introduce a method to utilize these powerful pre-trained transformers for a new downstream task, that is, automatic report generation for high resolution histopathology images.

\vspace{-0.4cm}
\section{Dataset}
\label{sec:dataset}

We get our imaging and associated text data from the Genotype-Tissue Expression (GTEx) portal\footnote{https://www.gtexportal.org/home/histologyPage}, same as \cite{zhang2020evaluating}. We divide our dataset into 23517 training, 603 validation and 1000 testing samples.
We do this because this is the only publicly available high resolution histology data with associated descriptions of each histology slide at the time of writing this paper that we could find. Data from Tsuneki and Kanavati \cite{tsuneki2022inference} comes in the form of 300x300 patches\footnote{https://zenodo.org/record/6021442} and the ARCH dataset by Gamper and Rajpoot\cite{gamper2021multiple} are of mixed quality, magnifications, and resolutions that do not meet our criteria of working with high resolution Whole Slide Images which are the most likely form of data available in health systems.

% \subsection{Caption pre-processing}
Data in the GTEx portal comes in a tabular format with $tissue\_type$, $sex$ and $pathology\_notes$ in separate columns. We create the description for each tissue in the following format:

\textit{this is a \{tissue\_type\} tissue from a \{sex\} patient and it has \{pathology\_notes\}.}

An example caption would look like this: 
\textit{this is a \textbf{small intestine - terminal ileum} tissue from a \textbf{male} patient and it has \textbf{6 pieces, prominent lymphoid component in 4 of 6 pieces.}}

\vspace{-0.6cm}
% This allows us to capture not only the pathological information provided, but also the $sex$ and $tissue\_type$
\section{Method}
\label{sec:majhead}

\subsection{WSI encoding using HIPT}
\label{sec:wsienc}

To encode the whole WSI for caption generation, we use the HIPT (Hierarchical Image Pyramid Transformer) architecture as described in \cite{chen2022scaling}, where the authors first describe the self-supervised pre-training of multiscale Vision Transformers (ViT) using the DINO method for knowledge distillation pioneered by \cite{caron2021emerging} on 10,678 WSIs. They then describe using this pre-trained multiscale ViT for downstream tasks like slide level classification, survival prediction and further using the unique attention maps generated by the ViTs for finding morphological phenotypes. They show that a HIPT based WSI level encoder outperforms current SOTA in multiple instance learning for histopathology classification, that is, CLAM-SB \cite{lu2021data}.

Here, we explain the steps taken to encode the WSI using HIPT as that will inform our use later when we use it for report generation. For $\mathrm{ViT_{L}}$-$l$, $L$ is the size of the patch and $l$ is the size of the $l \times l$ non overlapping tokens extracted from the $L \times L$ or $\textbf{x}_L$ image following the notation described in \cite{chen2022scaling}.

\begin{enumerate}
    
    \item Create $M$ 4096x4096 patches for each WSI, taking care that each patch contains more than 50\% tissue area.
    
    \item Initialize and freeze $\mathrm{ViT_{256}}$-16 and $\mathrm{ViT_{4096}}$-256 subnetworks. 

    \item For each $\textbf{x}_{4096}$ patch, first extract the $\mathrm{[CLS]}_{256} \in \mathbb{R}^{1\times384}$ representation for each 256x256 sized patch from $\mathrm{ViT_{256}}$-16. Since a $\textbf{x}_{4096}$ has 256 $\textbf{x}_{256}$ patches, each $\textbf{x}_{4096}$ can now be represented using a vector $\mathcal{P}_{256} \in \mathbb{R}^{256\times384}$.

    \item We also extract the $\mathrm{[CLS]}_{4096} \in \mathbb{R}^{1 \times 192}$ representation for each $\textbf{x}_{4096}$ patch from $\mathrm{ViT_{4096}}$-256.

    \item We concatenate the $\mathrm{[CLS]}_{4096}$ and $\mathcal{P}_{256}$ representations for each $M$-patch such that each WSI is now represented with a vector $\mathcal{P}_{WSI} \in \mathbb{R}^{M\times 256 \times 576}$.
    % \item For each $\textbf{x}_{4096}$ patch, first extract $\mathrm{[CLS]}_{256} \in \mathbb{R}^{1\times384}$ token vector from the $\mathrm{ViT_{256}}$-16, which is the summarizing token that encodes the 16x16 patch. And since we have 256 such 16x16 tokens per 256x256 patch, there are 256 such $\mathrm{[CLS]}_{256}$ tokens.
    % Since $\mathrm{ViT_{4096}}-256$ takes in patches of size (256x256), each patch is now represented by the [CLS] token.
    
    % \item The $\mathrm{[CLS]}_{256}$ token for each non overlapping $\textbf{x}_{256}$ patch from the $\textbf{x}_{4096}$ patch is extracted from the $\mathrm{ViT_{256}}$-16 and used to extract the $\mathrm{[CLS]}_{4096} \in \mathbb{R}^{1\times192}$ from $\mathrm{ViT_{4096}}$-256. This token now encodes the whole of $\textbf{x}_{4096}$ patch. Here, we also concatenate the mean of the 256 $\mathrm{[CLS]}_{256}$ to create $\mathrm{[CLS]}_{patch} \in \mathbb{R}^{1\times576}$ token, that is the final representation for each $\textbf{x}_{4096}$ patch.

    % \item At the end, for each WSI we have a representation vector, $\mathcal{P} \in \mathbb{R}^{M\times576}$ consisting of $M$ $\mathrm{[CLS]}_{patch}$ tokens that can now be used for caption generation.
\end{enumerate}

Chen et al. \cite{chen2022scaling} generously provide their pre-trained weights for $\mathrm{ViT_{256}}$-16 and $\mathrm{ViT_{4096}}$-256 in a GitHub repository\footnote{https://github.com/mahmoodlab/HIPT/} that we utilize to generate multi-scale representations of our Whole Slide Image.

\begin{table*}[htbp]
\centering
  \label{tab:results}
  {\caption{Caption Generation Results on test set; N prefix denotes performance on path. notes section; mean/std over 3 runs}}%
  {%
    \begin{tabular}{|l|l|l|l|l|l|l|l|l|l|}
    \hline
      \bfseries Model  & \bfseries Tiss. Acc.(\%)  & \bfseries BLEU-4 & \bfseries ROUGE-L  & \bfseries METEOR & \bfseries N BLEU-4 & \bfseries N ROUGE-L & \bfseries N METEOR\\
    \hline
   HIPT-BERT & \textbf{89.534$\pm$0.379} & \textbf{0.578$\pm$0.002} & 0.742$\pm$0.002 & \textbf{0.703$\pm$0.002} & \textbf{0.119$\pm$0.008} & \textbf{0.429$\pm$0.004} & 0.381$\pm$0.003 \\
   HIPT-BioBERT & 87.567$\pm$2.060 & 0.576$\pm$0.004 & 0.749$\pm$0.002 & 0.704$\pm$0.002 & 0.117$\pm$0.004 & 0.417$\pm$0.003 &0.370$\pm$0.007 \\
    % \hline
    HIPT-ClinicalBERT & 86.467$\pm$0.115 & 0.571$\pm$0.003 & \textbf{0.758$\pm$0.006} & 0.702$\pm$0.002 & 0.111$\pm$0.005 & 0.418$\pm$0.001 &\textbf{0.383$\pm$0.034} \\
    % HIPT-BioBERT & 87.567$\pm$2.060 & 0.576$\pm$0.004 & 0.749$\pm$0.002 & 0.704$\pm$0.002 & 0.117$\pm$0.004 & 0.417$\pm$0.003 &0.370$\pm$0.007 \\
    \hline
    \end{tabular}
    }
\end{table*}

\vspace{-0.2cm}

\begin{table*}[htbp]
\centering
  \label{tab:results2}
  {\caption{Effect of number of trained parameters; N prefix denotes performance on path. notes section; M=Million, mean/std over 3 runs}}%
  {%
    \begin{tabular}{|l|l|l|l|l|l|}
    \hline
      \bfseries Decoder Layers  & \bfseries \# Trained Params &\bfseries Tiss. Acc.(\%)  & \bfseries N BLEU-4 & \bfseries N ROUGE-L & \bfseries N METEOR \\
    \hline
   HIPT-BERT (last 1 layer+xattn) & 37.3M & 82.067$\pm$1.537 & 0.094$\pm$0.005 & 0.392$\pm$0.009 & 0.338$\pm$0.012\\
    % \hline
    HIPT-BERT (last 2 layer+xattn) & 44.4M & 85.767$\pm$2.902 &  0.115$\pm$0.002& 0.420$\pm$0.009 & 0.367$\pm$0.004\\

    HIPT-BERT (last 3 layer+xattn) & 51.5M & 89.234$\pm2.101$ & 0.124$\pm$0.005 &  0.426$\pm$0.005 & 0.384$\pm$0.006\\
    \hline
    \end{tabular}
    }
\end{table*}

\subsection{Caption Generation using BERT}
\label{sec:capgen}

Here, we go over the steps to train the VisionEncoderDecoder Model for caption generation also illustrated in Figure \ref{fig:overview}.

\begin{enumerate}
    % \item Generate a low resolution (1024x1024) thumbnail image and $M$-4096x4096 sized patches at 20x magnification from all available WSIs.
    
    % \item Use a pre-trained ResNet-18 to extract the thumbnail representation, $\mathcal{T} \in \mathbb{R}^{1\times512}$, and the pre-trained $\mathrm{ViT_{256}}$-16 and $\mathrm{ViT_{4096}}$-256 to extract patch based representation, $\mathcal{P} \in \mathbb{R}^{M\times576}$, as described in Section \ref{sec:wsienc}.
    \item We generated $\mathcal{P}_{WSI} \in \mathbb{R}^{M\times 256 \times 576}$ in Section \ref{sec:wsienc} that is able to represent all the $M$-patches for each WSI. We feed this in to a trainable Attention layer to create a weighted representation of size $\mathbb{R}^{1 \times 256 \times576}$.     
    \item We feed this multi-level representation  for each WSI into the ClinicalBioBERT based decoder following a similar process as \cite{li2023trocr}.
\end{enumerate}

\subsection{Training}
\label{sec:train}
Our training process first requires us to extract $\mathcal{P}_{WSI} \in \mathbb{R}^{M\times 256 \times 576}$ for each WSI. We then use these extracted representations in an end to end training process that first encodes the WSI using HIPT and generates weighted representations using the attention layer, which is then fed into BERT for language modeling (BertLMHeadModel)\footnote{https://huggingface.co/docs/transformers/model\_doc/bert}-based decoder. The Cross Entropy loss function is used for training.
We utilize PyTorch 1.12 and PyTorch Lightning 2.0.5 for our experiments (\textcolor{blue}{code available here}\footnote{https://github.com/ssen7/histo\_cap\_transformers}). We use a batch size of 1, with gradient accumulation of 16. We use a learning rate of 2e-5 with the Adam optimizer and train for 20 epochs for every model and use the model weights for the epoch for which the validation loss was the lowest. We also incorporate value based gradient clipping to tackle the exploding gradient problem. All our models were trained on a single NVIDIA A100 GPU, and mixed precision training helped reduce training time from 10hrs to 2hrs. Both pre-trained vision transformers, that is, $\mathrm{ViT_{256}}$-16 and $\mathrm{ViT_{4096}}$-256 were frozen during training and only the attention layer and non-linear projection layer with ReLU activation were trained. We unfroze different layers of the decoder to test which setup worked best according to our evaluation criteria.
% There are a number of possible combinations of encoders, fine-tuning targets and representations that can be tweaked to find the best possible combination of all. For example, we can choose to initialize ResNet-18 randomly or use ImageNet pre-trained weights. We can also choose to fine-tune the ResNet-18 block or freeze it during training. In the following section, we detail our experiments.

\vspace{-0.2cm}
\subsection{Evaluation}
\label{sec:eval}

We use natural language evaluation metrics like BLEU-4 \cite{papineni2002bleu}, METEOR \cite{banerjee2005meteor} and ROUGE-L\cite{lin2004rouge} scores to evaluate how closely the generated captions match the actual captions. We report these metrics for the overall sentence and for just the $pathology\_notes$ section. We also calculate the accuracy of generated $tissue\_type_g$ to the actual $tissue\_type_a$. These metrics, we believe, allow us to holistically evaluate our model. We report our results on a held out test set of 1000 patients. Note that we forgo reporting metrics on gender classification as there is no visual way to determine gender from tissue images and therefore it is a meaningless metric.
 
\section{Experiments}

% \subsection{Training Details}

\subsection{General BERT vs BioBERT vs BioClinicalBERT}
Our first experiment is designed to test the performance of general BERT (bert-base-cased), BioBERT and BioClinicalBERT to test the effect of domain specific pre-training on captioning performance. These pre-trained decoders are comparable because BioBERT and BioClincialBERT was initialized using BERT-base and then pretrained on PubMed research articles and MIMIC notes \cite{alsentzer2019publicly}. We freeze all layers of the encoder except for the attention layer and unfreeze all the layers of the BERT decoder, leading to 137M trainable parameters of 140M total parameters. During inference, we utilize greedy decoding (beam size of 1). The results can be seen in Table 1. All models were trained 3 times and the metrics reported are the average and standard deviation over those runs.

We can see that both BioBERT and BioClinicalBERT do not give use better results than general BERT on our evaluation metrics, even though the number of trained parameters remain the same (137M). We believe this is because we are unfreezing all layers of the decoder model, thereby re-training all weights regardless of initialization. This phenomenon is investigated in \cite{harrigian2023eye}, where the authors show that domain specific initialization of language models offer little to no benefit for their task of concept extraction on ophthalmology notes. The authors conclude that domain specific fine-tuning is more important than domain specific weight initialization, which we can confirm based on the results of Table 1. We believe the ROUGE-L score on the whole note is higher using BioClinicalBERT because of the artifacts of creating a template for our full note as detailed in Section \ref{sec:dataset}.
% Here we see that HIPT-BioClinicalBERT slightly outperforms the HIPT-BERT combination of encoder-decoder in the BLEU-4 score and Tissue accuracy criteria. Therefore, we use the HIPT-BioClinicalBERT combination for further experiments. This also validates our theory that domain specific pre-training helps in downstream tasks related to that particular domain, although very slightly in our case.
% We first train a baseline model using only the thumbnail images for the WSIs and the same method as \cite{xu2015show}. We experiment first with a randomly initialized ResNet-18 and then with initializing it with Imagenet pre-trained weights. We then use only the ViT based representation for encoding the WSI, to understand the contribution of the ViT. And finally, we combine the thumbnail representations with the WSI level representations from the ViT based encoder and use that to feed the LSTM based decoder. 

\vspace{-0.4cm}
\subsection{Fine-tuning different layers}

We sequentially unfreeze the last few layers of the BERT-base decoder along with unfreezing all the cross attention (denoted as xattn in the table) layers to test which setting provides the best performance for training the fewest parameters. The cross attention layers in the BERT decoder need to be trained from scratch since they do not have pre-trained weights. We unfroze the last $N$-layers where $N \in [1,2,3]$ of the 12 total layers available along with all the cross attention layers. These results are available in Table 2.

Of the total 140M trainable parameters, we can see that only training the cross attention layers and the last 3 layers of the BERT decoder (51.5 million parameters) achieves comparable results. This is helpful because it reduces training time as the number of parameters to optimize are lower.
\section{Conclusion}

In this paper, we show that powerful pre-trained ViT based representations could be used to encode a very high resolution histology image slide for another downstream task, that is, successful automatic report generation. We have also validated that self supervised pre-training is helpful, as it lessens the burden of training data hungry models from scratch every time. We also find that domain specific initialization of the language modeling decoder does not provide meaningful performance gain, therefore leading credence to the theory that domain specific fine tuning achieves better results than pre-trained weight initialization.
% Most previous methods for caption generation in histopathology required having a method to encode these high resolution images into manageable representations that could then be used for model learning, which can now successfully be replaced by powerful self-supervised pre-trained encoders.

Of the 40 different tissue types available in our test data, our model was able to correctly classify over 89\% of them, which suggests that we can successfully use the captioning model for multi-class classification, which is a valuable objective. 
% Additionally, we can look at caption generation as a self-supervised objective that allows us to combine images and text based data. Data like radiology and histopathology slides and associated text reports are stored across hospital systems and can be used for self-supervised learning and do not require trained clinical professionals to assign labels that is often necessary for most supervised learning tasks. Indeed, \cite{lu2023visual} and \cite{gamper2021multiple} propose similar objectives, that is, combining vision and language data to pre-train powerful encoders for histopathology that can then be used for zero-shot learning and produce promising results. 

% In the same vein, we can look at the caption generation as a way to incorporate text based data into these already powerful representations generated by the vision transformers trained by \cite{chen2022scaling}. 
We present, in this paper, a method to utilize powerful pre-trained transformer models for automatic report generation specifically for histopathology with an end-to-end training mechanism that to the best of our knowledge had not been proposed before. 
% Transformer-based language models significantly outperform RNN-based models, as BioClinicalBERT being pre-trained on clinical notes and medical research articles encodes complex information that the RNN-based encoders used in previous papers on automatic image captioning for histopathology cannot. 
We believe this work broadens the scope of research in histopathology by introducing transformers in place of traditional CNN-RNN based encoder-decoder models for caption generation.

\vspace{-0.4cm}
\section{Acknowledgments}
\label{sec:acknowledgments}
% \section*{Funding Statement}
The work in the paper was partially supported by  the National Center for Advancing Translational Science of the National Institutes of Health Award UL1TR003015/ KL2TR003016.

\vspace{-0.4cm}
\section{Compliance with ethical standards}
\label{sec:ethics}
This research study was conducted retrospectively using human subject data made available in open access available here\footnote{https://www.gtexportal.org/home/histologyPage}. Ethical approval was not required as confirmed by the license attached with the open access data.

\bibliographystyle{IEEEbib}
\bibliography{strings,refs}

\end{document}